\documentclass[trans]{IEEEtran}
\IEEEoverridecommandlockouts
\usepackage{cite}
\usepackage{amsmath,amssymb,amsfonts}
\usepackage{algorithmic}
\usepackage{graphicx}
\usepackage{textcomp}
\usepackage{xcolor}
\usepackage{booktabs}
\usepackage{multirow}
\usepackage{subfigure}
\usepackage{url}
\usepackage[ruled,linesnumbered]{algorithm2e}
\usepackage[colorlinks,urlcolor=blue,linkcolor=blue,citecolor=blue]{hyperref}

\usepackage{color,array}

\def\BibTeX{{\rm B\kern-.05em{\sc i\kern-.025em b}\kern-.08em
    T\kern-.1667em\lower.7ex\hbox{E}\kern-.125emX}}
\begin{document}

\title{Learning Resilient Formation Control of Drones with Graph Attention Network\\
\thanks{This work has been submitted to the IEEE for possible publication. Copyright may be transferred without notice, after which this version may no longer be accessible. J. Xiao, Q. Jia, and M. Feroskhan are with the School of Mechanical and Aerospace Engineering, Nanyang Technological University; X. Fang is with the School of Electrical and Electronic Engineering, Nanyang Technological University, Singapore, 639798, Singapore. (e-mail: jiaping001@e.ntu.edu.sg; fa0001xu@e.ntu.edu.sg; jiaql@mail.nwpu.edu.cn; mir.feroskhan@ntu.edu.sg)}
}

\author{\IEEEauthorblockN{Jiaping~Xiao,~Xu~Fang,~Qianlei~Jia,~Mir~Feroskhan}}

\maketitle

\begin{abstract}

The rapid advancement of drone technology has significantly impacted various sectors, including search and rescue, environmental surveillance, and industrial inspection. Multidrone systems offer notable advantages such as enhanced efficiency, scalability, and redundancy over single-drone operations. Despite these benefits, ensuring resilient formation control in dynamic and adversarial environments, such as under communication loss or cyberattacks, remains a significant challenge. Classical approaches to resilient formation control, while effective in certain scenarios, often struggle with complex modeling and the curse of dimensionality, particularly as the number of agents increases. This paper proposes a novel, learning-based formation control for enhancing the adaptability and resilience of multidrone formations using graph attention networks (GATs). By leveraging GAT's dynamic capabilities to extract internode relationships based on the attention mechanism, this GAT-based formation controller significantly improves the robustness of drone formations against various threats, such as Denial of Service (DoS) attacks. Our approach not only improves formation performance in normal conditions but also ensures the resilience of multidrone systems in variable and adversarial environments. Extensive simulation results demonstrate the superior performance of our method over baseline formation controllers. Furthermore, the physical experiments validate the effectiveness of the trained control policy in real-world flights.

\end{abstract}


\section{Introduction}

The emergence of drone technology has significantly impacted various sectors, including search and rescue, environmental surveillance, and industrial inspection. Multidrone systems, in particular, offer distinct advantages such as enhanced efficiency, scalability, and redundancy compared to single-drone operations \cite{xiao2022collaborative, zhou2023racer, xiao2024toward}. To fully exploit these advantages, drones typically need to collaborate with each other and maintain desired formations \cite{jinhu2024}, which necessitates reliable inter-agent communication and egocentric observations during flight. Formation control is pivotal for the operational effectiveness of coordinating multiple agents to move cohesively while maintaining a desired geometric arrangement throughout the mission \cite{fang2021distributed}. Traditional formation control strategies often rely on predetermined formation configurations \cite{du2019distributed} and dependable communication or measurements, such as distance and relative position \cite{4653105}, primarily focusing on stability and consensus issues. The consensus-based approach, foundational to formation control, is thoroughly examined in \cite{olfati2007consensus, su2016distributed, du2019distributed, liu2020fixed, yang2022data, fang2023distributed}, providing key insights into leader-follower and distributed coordination. While the leader-follower consensus formation benefits from simplified coordination and faster decision-making due to the presence of a central leader, it may be less resilient to failures in the central leader compared to distributed consensus formation, where all agents equally contribute to the decision-making process, leading to a more decentralized and robust coordination. For instance, Fang et al. \cite{fang2023distributed} offer an in-depth analysis of distributed algorithms for robust formation control and maneuvering, addressing challenges in dynamic environments. From the measurement constraint perspective, existing formation control strategies have been developed into four main types: displacement-based \cite{ren2007distributed}, distance-based \cite{baillieul2003information}, bearing-based \cite{zhao2015translational}, and angle-based \cite{10040975}. To mitigate the insufficiencies of these single constraint-based formation control methods, hybrid constraint-based approaches, such as distance-angle-based formation control \cite{shao2024directed}, have been developed. However, these methods primarily focus on reducing measurement costs while neglecting communication reliance and obstacle avoidance, which can result in unpredictable performance under communication loss.

The unpredictable nature of real-world scenarios demands a more flexible and resilient formation approach, especially when standard communication channels, such as WiFi, Bluetooth, and UWB, are disrupted by environmental complexity or cyberattacks such as false data injection attacks and denial-of-service (DoS) attacks \cite{xiao2022cyber}. This need is particularly pronounced in missions where maintaining cohesive formations is crucial, such as search and rescue operations in cluttered environments. The resilience of drone formations in adversarial or uncertain environments is critical for maintaining operational integrity. To address this need, many classical defense solutions have been proposed to ensure the resilience of multiagent systems via attack prevention, attack detection, and attack resilience \cite{pirani2023graph}. Pasqualetti et al. \cite{pasqualetti2011consensus} discuss the vulnerabilities and defense mechanisms in networked systems, providing a framework for understanding resilience in multiagent formations. Furthermore, Shames et al. \cite{shames2011distributed} delve into resilient networked control, offering strategies to counteract the effects of potential cyberattacks. These methods generally consider specific system dynamics and are difficult to extend to large-scale formation missions. Graph theory has been widely used to model and analyze the resilience of distributed multiagent systems \cite{Sarrafan2024resilient, li2023resilient, pirani2023graph, han2024distributed} due to its inherent advantage in describing large-scale connected systems. For instance, a resilient consensus control method with robust sliding mode and sampled neighboring information is proposed in \cite{li2023resilient} to handle external disturbances and DoS attacks simultaneously. In \cite{han2024distributed}, a distributed adaptive consensus control with attack detection and a switching strategy is developed to actively mitigate attack effects. Meanwhile, event-based control methods \cite{chen10323259, han2023distributed} have been introduced to reduce the communication burden in continuously monitoring the status of agents. However, these traditional consensus-based algorithms and event-triggered methods that rely on graph theory have fundamental limitations in detecting attacks (\!\!\cite{pirani2023graph}, Remark 1), where attack periods are assumed to be known, and involve complex computation and subgraph switching strategies for large-scale systems, which are not practical in real-world scenarios.

Recent developments in deep learning have shown promising solutions to address the computation overhead in formation control by learning the model dynamics and information exchange \cite{tolstaya2020learning, zhao2021robust, sui2021formation, zhao2022data, yu2023neural, pu2023deep}. Tolstaya et al. \cite{tolstaya2020learning} explore the use of graph neural networks for decentralized control in multiagent systems, showcasing the potential of learning-based approaches to enhance formation control. In \cite{ zhao2021robust, yu2023neural}, model-free robust formation controllers are trained with reinforcement learning for quadrotor UAVs without requiring knowledge of the dynamics. In \cite{sui2021formation, pu2023deep}, model-guided formation control with collision avoidance methods are developed through DRL for mobile robots in a leader-follower structure. However, achieving resilient formation control for nonlinear multidrone systems under adversarial attacks is much more challenging and has not been fully explored. To address these challenges, this work proposes a lightweight, model-free, resilient formation control solution for multidrone systems, integrating Graph Attention Networks (GATs) and deep reinforcement learning (DRL) into the formation control problem. GATs have revolutionized the way information is processed in graph-structured data. The initial introduction of GATs by Veličković et al. \cite{velivckovic2017graph} has led to extensive research exploring their applications in various domains. For instance, Lee et al. \cite{lee2019graph} demonstrate the effectiveness of attention mechanisms in graph neural networks for dynamic node representation learning, which is directly applicable for understanding complex agent interactions in multidrone systems. GATs uniquely adapt internode relationships based on attention scores from neighbors, allowing the extraction of implicit state features to enhance the adaptability and resilience of drone formations. By considering the control cost and collision rewards, we find that, by utilizing the dynamic capabilities of GATs, drone formations can maintain effective communication and optimal formation structures with fewer collisions in different formation maneuvers, even under DoS attacks. The key contributions of this work are:

1) To eliminate the need for attack detection and subgraph switching in traditional consensus-based methods \cite{li2023resilient, han2024distributed} and event-triggered methods \cite{han2023distributed}, this paper proposes a novel learning-based method without requiring knowledge of complex model dynamics for multidrone resilient formation control using GATs, designed to enhance adaptability and resilience against DoS attacks.

2) To reduce communication costs, we design a dual-mode formation control framework integrating leader-follower formation and distributed formation, leveraging the adaptive switching capability of GATs to handle variable observations.

3) We further validate the effectiveness and generalization capability of the trained resilient formation control policy in real-world flight with zero-shot Sim2Real to a drone swarm.

The rest of this paper is organized as follows: Section II describes some preliminary works. Section III elaborates on the proposed methodology. Section IV presents the experimental results and discussions. Section V concludes this paper.


\section{Preliminary}
\subsection{Graph Theory}
Graph theory involves using graph structure to model pairwise relations between $N$ objects labeled from $1$ to $N$. A weighted graph is defined as a pair \(G =(V,E,W)\) where \(V\) is a set of $N$ nodes and \(E\subseteq V \times V\) is a set of edges that connect pairs of nodes. A direct edge \(e_{ij} \! = \! (i, j) \! \in \! E\) links vertex \(v_i\) to vertex \(v_j\) ($i$, $j$ for simplicity). $W=[w_{ij}] \in \mathbb{R}^{N\times N}$ is the associated adjacency matrix, where the element $w_{ij}\ge 0$ indicates the connection of an edge $(i,j)\in E$. The adjacency matrix is defined as:
\begin{equation}
    W_{ij} = \begin{cases} 
w_{ij}\neq 0 & \text{if there is an edge between } i \text{ and } j, \\
0 & \text{otherwise}.
\end{cases}
\end{equation}

\begin{figure}
    \centering
    \includegraphics[width=3.3in]{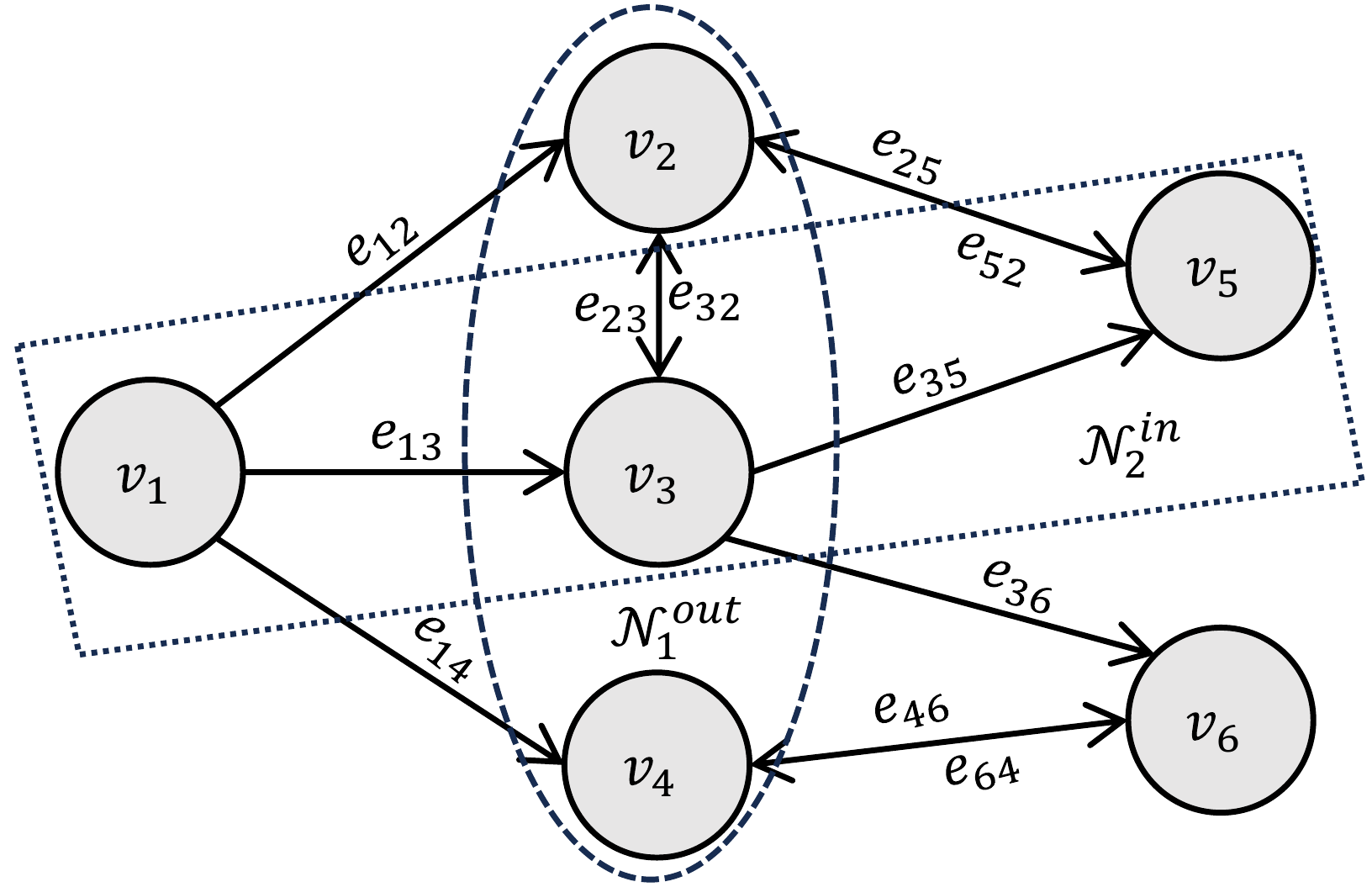}
    \caption{Illustration of one direct graph definition.}
    \label{fig:graph}
\end{figure}

For a directed graph, the adjacency matrix is asymmetric, i.e., $w_{ij} \!\neq\! w_{ji}$ while for an undirected graph, $w_{ij} \!=\! w_{ji}$. As illustrated in Fig. \ref{fig:graph}, the in-neighbors and out-neighbors of vertex $i$ are denoted as $\mathcal{N}_i^{in} := \{j\in V \mid (j,i) \in E, j\neq i\}$ and $\mathcal{N}_i^{out} := \{j\in V \mid (i,j) \in E, j\neq i\}$, respectively. The neighbors of vertex $i$ are $\mathcal{N}_i = \mathcal{N}_i^{in} = \mathcal{N}_i^{out}$ if the graph is undirected. A directed path from node $i$ to node $j$ is a sequence of ordered edges of the form $\{(i, i+1), \dots, (j-1, j)\}\subseteq E$.
The weighted in-degree of an agent $i$ is $d_i = \sum_{j\in \mathcal{N}_i^{in}} w_{ij}$. The graph Laplacian is denoted as \(L = D - W\), where $D = diag(d_i)\in \mathbb{R}^{N\times N}$ is the diagonal degree matrix. The Laplacian matrix plays a crucial role in studying graph properties and is extensively used in determining the connectivity of the graph. $G$ has a spanning tree if a node can be connected to all other nodes with available directed paths.

In the context of multiagent systems, nodes often represent agents, and edges represent the communication or interaction pathways between them. For a formation control task, the weighted graph $G$ can be particularly informative, indicating if agents are within communication range of each other to obtain corresponding measurements such as distance $d_{ij}$ or relative position $\boldsymbol{p}_{ij}$, and thus can coordinate for formation adjustments. The graph Laplacian 
$L$ is instrumental in analyzing and designing consensus-based control strategies, ensuring that the formation achieves desired configurations and adapts to changes in the environment or network.

\subsection{Graph Attention Network}
GATs \cite{velivckovic2017graph} extend the concept of neural networks to graph-structured data. They leverage the attention mechanism to weigh the influence of nodes on one another dynamically. The key idea is to allow nodes to extract their neighborhoods' features selectively, enabling the model to prioritize information from more important neighbors. Let's consider a graph \(G = (V, E, W)\) and a node \(i\). In GAT, the attention mechanism computes the coefficients that signify the importance of node \(j\)'s features to node \(i\) where $j \in \mathcal{N}_i$, with coefficients $\alpha_{ji}$ as:
\begin{equation}
\alpha_{ji} = \frac{\exp(\text{LeakyReLU}(\mathbf{a}^T[\mathbf{W}_q h_i \| \mathbf{W}_k h_j]))}{\sum_{j \in \mathcal{N}_i} \exp(\text{LeakyReLU}(\mathbf{a}^T[\mathbf{W}_q h_i \| \mathbf{W}_k h_j]))} 
\end{equation}
where \(h_i\) is the state vector of node \(i\), \(\mathbf{W}_q\) and \(\mathbf{W}_k\) are learnable weight matrices to encode the states of node $i$ and neighbour $j$ respectively, and \(\|\) denotes concatenation of two feature vectors, i.e., cascading into one feature vector. \(\mathbf{a}\) is a learnable weight vector to encode the concatenation of encoded states and \(\mathcal{N}_i\) represents the in-neighbors of node \(i\) for simplicity. The coefficients \(\alpha_{ji}\) are normalized across all choices of \(j\) using the softmax function. The LeakyReLU function is defined as: \(\text{LeakyReLU}(x) = x ~\text{if}~ x \ge 0~\text{and}~
\text{LeakyReLU}(x) = bx ~\text{otherwise}, \) where $b$ is a small negative constant (default -0.01 in PyTorch\footnote{https://pytorch.org/docs/stable/generated/torch.nn.LeakyReLU.html}). After calculating the attention coefficients, the node features are updated by aggregating the neighbors' features, weighted by the attention coefficients as

\begin{equation}
h_i' = \sigma\left(\sum_{j \in \mathcal{N}_i} \alpha_{ji} \mathbf{W}_v h_j\right)
\end{equation}
where \(\sigma\) is a non-linear activation function such as softmax or sigmoid, $\mathbf{W}_v$ is another learnable encoding weight matrix, and \(h_i'\) is the updated feature vector of node \(i\). This process can be stacked in multiple layers to enable deeper feature extraction. GATs have shown to be particularly effective in tasks where the structure of the graph is vital for the learning task, such as in the context of multidrone systems where the communication topology can influence the system's behavior significantly.

\subsection{Quadrotor Drone Dynamics}
In this work, quadrotor drones are used to conduct the formation maneuvering. The quadrotor drone dynamics can be modeled as a rigid body controlled by thrust and torque commands, which influence its linear and angular accelerations. The quadrotor drone dynamics can be represented by:
\begin{align}
    &m\ddot{\boldsymbol{p}} + m\boldsymbol{g}= \boldsymbol{f}_{\text{thrust}} \label{eq:pos_dyn}\\
       &{\boldsymbol{\dot q}}{}={} \frac{1}{2}\boldsymbol{\Omega}(\boldsymbol{\omega}) \boldsymbol{q} \\
    &I\dot{\boldsymbol{\omega}} + \boldsymbol{\omega} \times (I\boldsymbol{\omega}) = \boldsymbol{\tau}
\end{align}
where \(m\) is the mass of the drone, $\boldsymbol{p}$ is position of drone in the world frame. $\boldsymbol{g} = [0, 0, g]^T$ is the gravity of Earth. The \(\boldsymbol{f}_{\text{thrust}} = [c_x,c_y,c_z]^T\) is the thrust force vector generated by the rotors and projected into the world frame from the body frame with the rotation matrix $\boldsymbol{R}_{\boldsymbol{q}}$ and $\boldsymbol{q} = [q{_0},q_1,q_2,q_3]^T$ is the unit quaternion to describe the quadrotor's attitude. $\boldsymbol{\Omega}(\boldsymbol{\omega}_B)$ is the skew-symmetric matrix to describe the rotational dynamics. \(I\) is the inertia matrix, \( \boldsymbol{\omega} = [\omega_x, \omega_y, \omega_z]^T\) is the angular velocity vector, and \(\boldsymbol{\tau}\) represents the torques generated by the drone. Since formation is generally determined by the position of drones, we only consider the position dynamics in (\ref{eq:pos_dyn}) and the rotational dynamics is tracked with a high-bandwidth PID controller. We use the thrust vector $\boldsymbol{f}_{\text{thrust}}$ as the action vector $\boldsymbol{u}=[c_x, c_y, c_z]$, and $x=[\boldsymbol{p}, \dot{\boldsymbol{p}}]$ as the state vector.

\section{Methodology}
In this section, we first formulate a leader-follower formation control problem in normal condition and a distributed resilient formation control problem under DoS attack. We then adopt a GAT to represent the dynamic formation. We lastly train the GAT and the control policy with DRL by specifically designing the action space, observation space, and reward functions to obtain a distributed resilient formation control policy under DoS attack.

\begin{figure}
    \centering
    \includegraphics[width=0.9\linewidth]{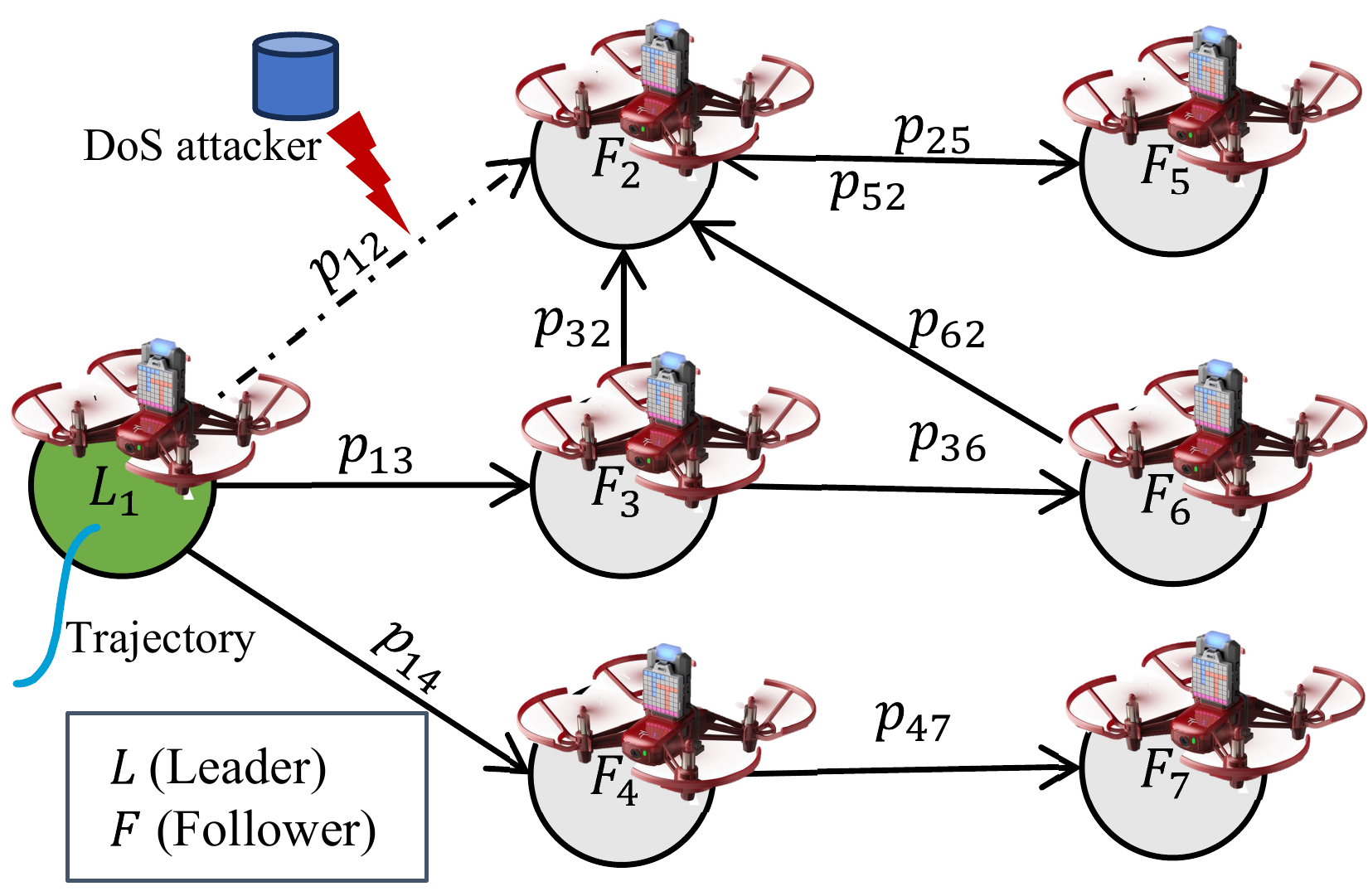}
    \caption{The drone formation with the DoS attacker.}
    \label{fig:droneformation}
\end{figure}

\subsection{Problem Formulation}
The objective of leader-follower formation control is to find a control law $\boldsymbol{u}_i$ for each agent $i\in \{1, 2, \cdots, N\}$ where $N$ is the number of agents, $i=1$ is the leader $L_1$ and others are the followers $F_k, k\in[2,N]$, such that

\begin{align}
&\lim\nolimits_{t\to\infty}\boldsymbol{e}_1(t)=\lim\nolimits_{t\to\infty}(\boldsymbol{p}_1(t)-\boldsymbol{p}_1^*(t))=0\label{formationobj1} \\ \notag 
&\lim\nolimits_{t\to\infty}\boldsymbol{e}_2(t)=\lim\nolimits_{t\to\infty}\left(\boldsymbol{p}_2(t)-\boldsymbol{p}_1(t)-\boldsymbol{p}^*_{12}\right)=0\\
\notag &...\\ 
&\lim\nolimits_{t\to\infty}\boldsymbol{e}_i(t)=\lim\nolimits_{t\to\infty}\left(\boldsymbol{p}_i(t)-\boldsymbol{p}_1(t)-\boldsymbol{p}^*_{1i}\right)=0\label{eq:formationobj}
\end{align}
where $\boldsymbol{p}_1^*(t) \in \mathbb{R}^3$ is the desired time-varying position for the leader $1$ and $\boldsymbol{p}^*_{ij} \in \mathbb{R}^3$ is the desired constant relative position from agent $i$ to agent $j$. For formation maneuvering, the leader is always required to track a desired time-varying trajectory $\{\boldsymbol{p}^*_1(t)\}$. 
Assuming that the state of the leader is broadcast to all other agents through the communication network, to achieve the objective of aforementioned formation control, we can design the control law $\boldsymbol{u}_i$ for each agent $i$ as follows:

\begin{align} 
&\boldsymbol{u}_1=m_1\left[\boldsymbol{g} \!+\! {\ddot{\boldsymbol{p}}_1^*(t)} \!-\! k_v(\dot{ \boldsymbol{p}}_1(t) \!-\! \dot{\boldsymbol{p}}_1^*(t)) \!-\! k_p(\boldsymbol{p}_1(t) \!-\! \boldsymbol{p}_1^*(t))\right] \notag \\
&\boldsymbol{u}_2=m_2\left[\ddot{\boldsymbol{p}}_1(t) +\boldsymbol{g}\!-\! k_v(\dot{\boldsymbol{p}}_{2}(t)-\dot{\boldsymbol{p}}_1(t))-k_p(\boldsymbol{p}_{12}(t)-\boldsymbol{p}^*_{12})\right] \notag \\
&... \notag \\
&\boldsymbol{u}_i=m_i\left[\ddot{\boldsymbol{p}}_1(t)+\boldsymbol{g}-k_v(\dot{\boldsymbol{p}}_{i}(t)-\dot{\boldsymbol{p}}_1(t))-k_p(\boldsymbol{p}_{1i}(t)-\boldsymbol{p}^*_{1i})\right] \label{eq:controller3}
\end{align}
where $\boldsymbol{p}_{1i}(t) = \boldsymbol{p}_i(t) - \boldsymbol{p}_1(t) \in \mathbb{R}^3$ is the relative position from the leader $L_1$ to the follower $F_i$, and $k_v$ and $k_p$ are positive controller gains. This formation control law is based on relative position measurements and has proved to be global stability \cite{10040975}. 
However, when communication to the leader is lost or DoS attack occurs for communication link $(1,a)$ where $a \in \{2,\cdots,N\}$, the relative position $\boldsymbol{p}_{1a}(t)$ is not available, which results the formation being unstable. In this work, we assume that $\ddot{\boldsymbol{p}}_1(t)\sim 0$ and the DoS attack is launched from a ground attacker at a fixed position $\boldsymbol{p}_{DoS}$ and set $\boldsymbol{p}_{1a}(t) = [0,0,0]^T$ when the distance $\|\boldsymbol{p}_a(t) - \boldsymbol{p}_{DoS}\| \leq \kappa$ where $\kappa$ is the attack range, i.e.,
\begin{equation} \label{eq:dos}
\setlength{\nulldelimiterspace}{0pt}
{\boldsymbol{p}_{1a}(t)} = \left\{ \begin{array}{l}
{\boldsymbol{p}_a(t) - \boldsymbol{p}_1(t)},\quad \text{if\;no\;DoS\;attack,}\\
 {[0,0,0]^T}, \quad \text{if}\;\text{DoS}\;\text{attack}\ (\|\boldsymbol{p}_a - \boldsymbol{p}_{DoS}\| \leq \kappa).
\end{array} \right.
\end{equation}

To achieve distributed resilient formation control under the DoS attack, one solution is to retrieve the relative position estimate $\hat{\boldsymbol{p}}_{1a}(t)$ from neighbours when the original communication is not available as illustrated in Fig. \ref{fig:droneformation}. Hence the distributed resilient formation control problem in this work is to design a control law $\boldsymbol{u}_a$ for the agent $a$ without communication with the leader to learn estimates of $\dot{\boldsymbol{p}}_{1}(t)$ and $\boldsymbol{p}_{1a}(t)$ from local neighbors, i.e.,  $\left[\hat{\dot{\boldsymbol{p}}}_{1}, \hat{\boldsymbol{p}}_{1a}\right] = h_i(\mathbf{W}\boldsymbol{p}_{ja}\|\mathbf{W}\dot{\boldsymbol{p}}_{j} \mid j\in \mathcal{N}_a)$ such that
\begin{equation}
    \boldsymbol{u}_a = m_a\left[\boldsymbol{g} - k_v(\dot{\boldsymbol{p}}_a(t)-\hat{\dot{\boldsymbol{p}}}_1(t))-k_p(\hat{\boldsymbol{p}}_{1a}(t)-\boldsymbol{p}^*_{1a})\right] + w_c \label{eq:ua}
\end{equation}
where $\mathbf{W}$ is the learnable weight matrix and $w_c$ is the learnable uncertainty item. Note that the $w_c$ is to compensate the uncertainties in the modeling. The neighbors of agent $a$ are those within the communication range $d_c$. To simplify the training process, we directly train one control policy $\boldsymbol{u}_a \sim \pi(\boldsymbol{u}_a \mid \boldsymbol{s}_a, \{\boldsymbol{s}_j\}_{j\in\mathcal{N}_a}, \mathbf{W}, w_c)$, where $\boldsymbol{s}_a$ and $\boldsymbol{s}_j$ are the states of the agent $a$ and its neighbors, respectively.

\subsection{Formation Representing with Graph Attention Network}

\begin{figure}
    \centering
    \includegraphics[width=0.85\linewidth]{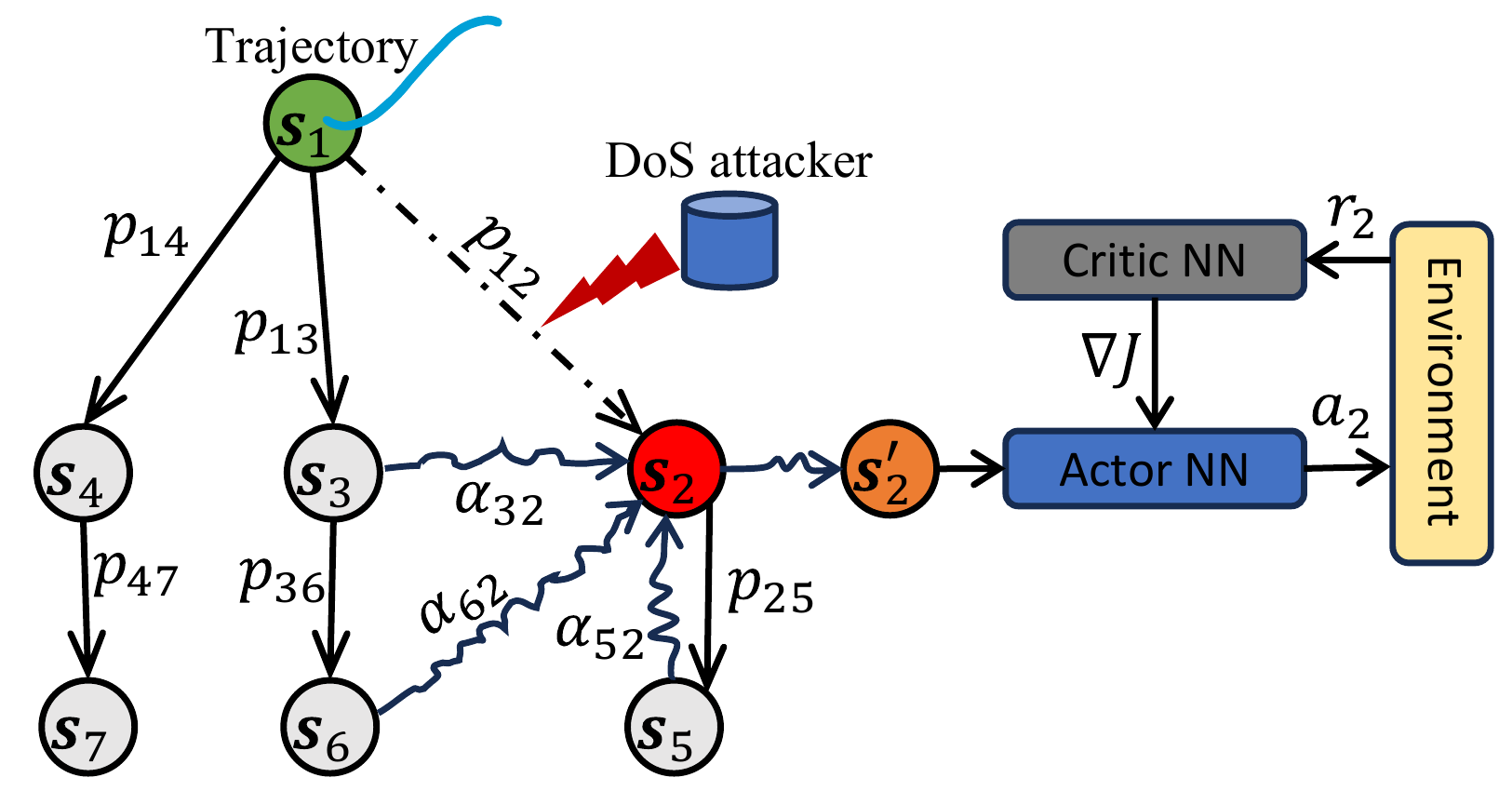}
    \caption{The overall architecture of the proposed distributed resilient formation control with GAT. The observable state of the attacked agent $\boldsymbol{s}'_2$ is extracted from a GAT with neighbour agents' states. If the agent 2 is within the attack range, the information $\boldsymbol{p}_{12}$ is not available. The RL is trained with the extracted state representation $\boldsymbol{s}'_2$.}
    \label{fig:resform}
\end{figure}

\begin{figure}
    \centering
    \includegraphics[width=0.85\linewidth]{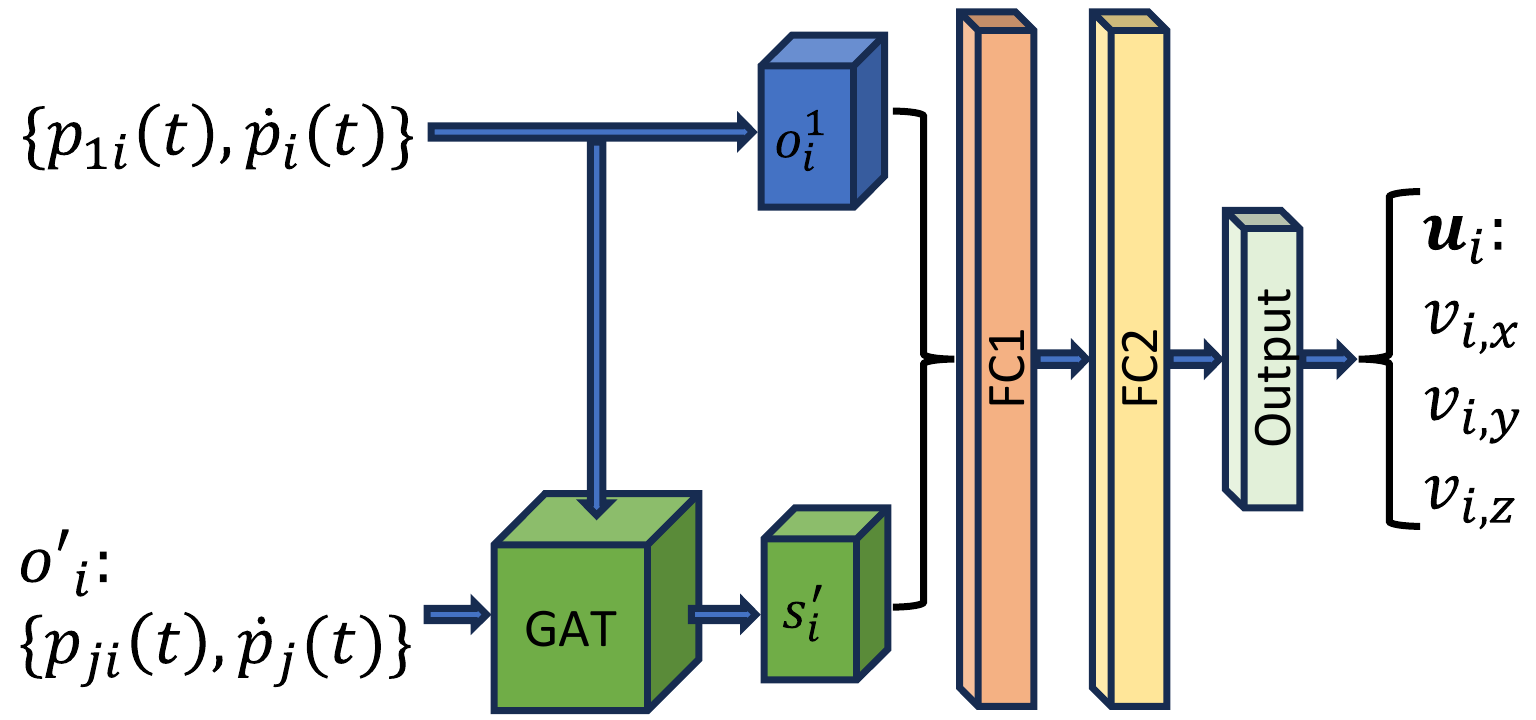}
    \caption{The overall architecture of the designed policy NN.}
    \label{fig:nn-arch}
\end{figure}

We define the graph representation of the drone formation as \(G = (V, E, W)\), where \(V\) denotes the set of drones and \(E\) represents the communication links. Each drone \(i\) has a state vector \(\boldsymbol{s}_i \in \mathbb{R}^6\) encapsulating its position and velocity. In the GAT, the attention mechanism computes the weights \(\alpha_{ji}\) which the drone \(i\) assigns to the states from drone \(j\):
\begin{equation}
    \alpha_{ji} = \frac{\exp\left(\text{LeakyReLU}\left(\mathbf{a}^T[\mathbf{W}_q\boldsymbol{s}_i \parallel \mathbf{W}_k\boldsymbol{s}_j]\right)\right)}{\sum_{j \in \mathcal{N}_i} \exp\left(\text{LeakyReLU}\left(\mathbf{a}^T[\mathbf{W}_q\boldsymbol{s}_i \parallel \mathbf{W}_k\boldsymbol{s}_j]\right)\right)}
\end{equation}
where \(\mathcal{N}_i\) denotes the neighbors of drone \(i\). The estimated state \(\boldsymbol{s}_i'\) after applying the attention mechanism is:
\begin{equation}
    \boldsymbol{s}_i' = \sigma\left(\sum_{j \in \mathcal{N}_i} \alpha_{ji} \mathbf{W}_v\boldsymbol{s}_j\right)
\end{equation}
where \(\sigma\) is a sigmoid activation function here. In such way, the attacked agent $a$ can obtain the leader state estimates from neighbours. Note that the leader state estimates from the GAT are learned with reinforcement learning by designing suitable reward functions in this study. The GAT is used to encode state representations from the input information of neighbours and then is cascaded with the control policy neural network (NN). The overall training structure and the policy NN architecture are illustrated in Fig. \ref{fig:resform} and Fig. \ref{fig:nn-arch}, respectively.

\begin{figure}
    \centering
    \includegraphics[width=0.85\linewidth]{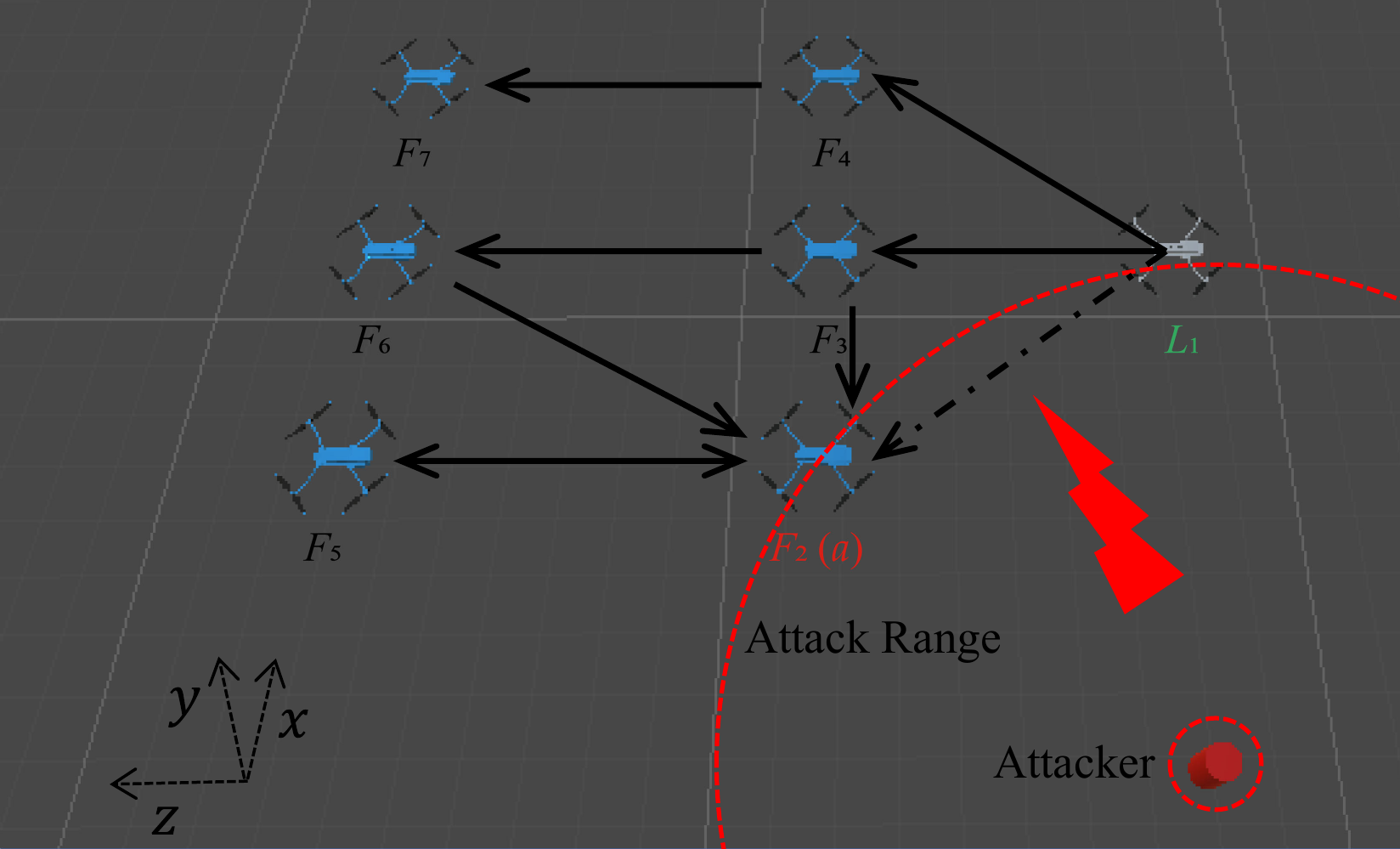}
    \caption{The simulation environment developed in Unity. The information $\boldsymbol{p}_{1a}$ of adversarial follower $F_2$ is blocked if it is within the attack range.}
    \label{fig:sim}
\end{figure}

\subsection{Learning Dual-Mode Resilient Formation Control Policy}
To save the communication costs, a dual-mode resilient formation control policy is adopted. In the normal condition where the communication to the leader is accessible, the policy is trained with the observation from the leader's state while the inputs of the attention module is padded with zeros. Under the DoS attack, the attention module is adopted to extract the features from the variable number of neighbours' state. The control policy is learned through DRL, which is generally modeled as a Markov decision process $M := \langle S, O, A, R, P, \gamma \rangle$ to maximize the expected cumulative reward from interactions with the operating environment based on observations. At each step, the agent choose an action $a \in A$ based on the local observation $o \in O$ from the global state $S$. This agent influences the environment with state transition as per the function $P$ and receives an instant reward $r \in R$. The discount factor $\gamma$ determines the importance of future rewards in an episode. The operating environment, state and observation space, action space, reward functions and training scheme are elaborated in the following.

\subsubsection{Environment}
\begin{figure}
    \centering
    \includegraphics[width=0.85\linewidth]{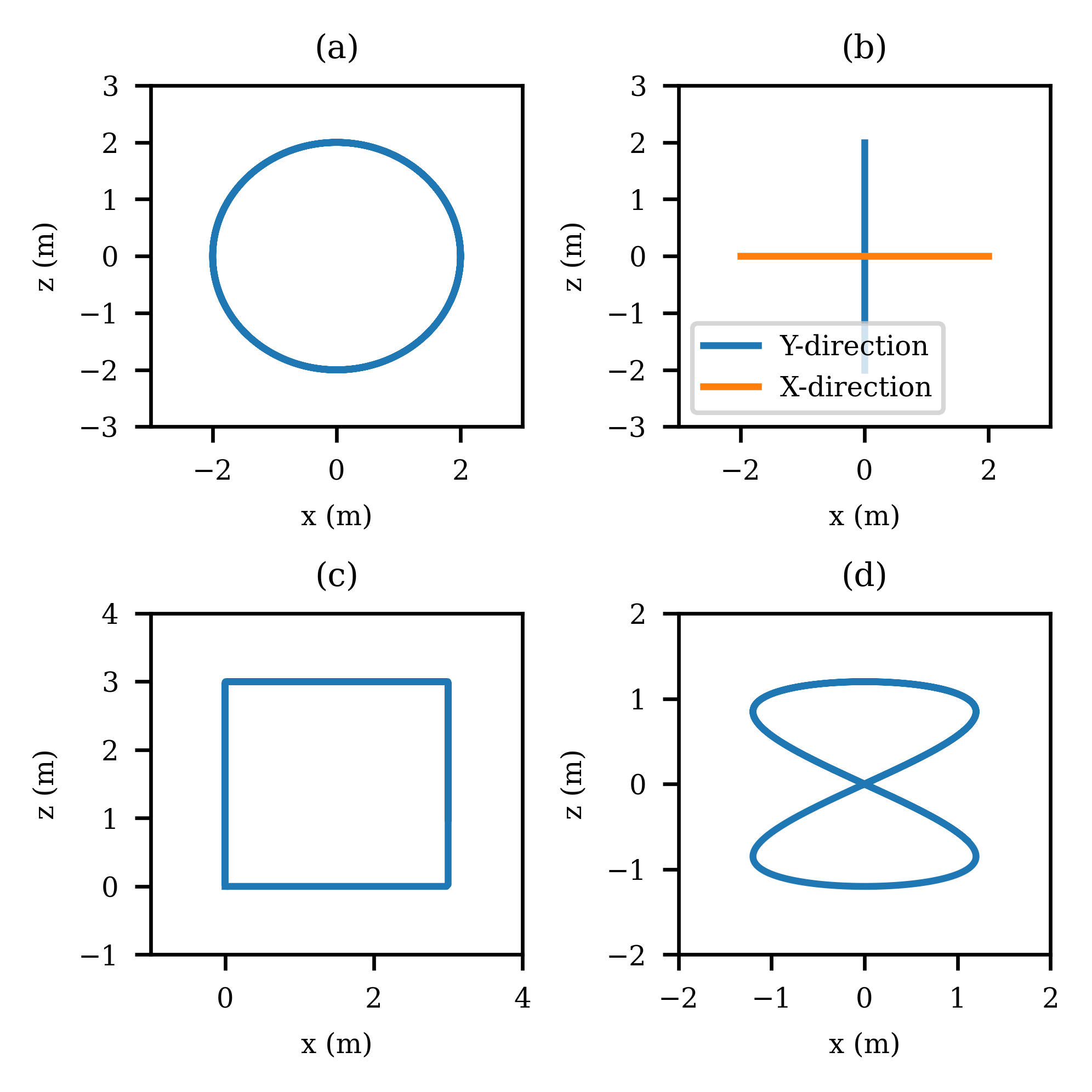}
    \caption{The trajectory pool includes different trajectories for the leader agent $L_1$ to track in each training episode. (a)Circular trajectory with $[x,z] = 2\cdot[\cos(t), \sin(t)]$; (b) Straightline trajectory with length of $4m$; (c) Square trajectory with length of $3m$; (d) Figure-eight trajectory with $[x,z]=1.2\cdot[\sin(t), \cos(t/2)]$.}
    \label{fig:trajpool}
\end{figure}

The training environment is developed with a Unity simulation engine, as illustrated in Fig. \ref{fig:sim}. The leader-follower formation framework is adopted in simulation, where the leader track a trajectory $\boldsymbol{\tau}^* = \{\boldsymbol{p}_1^*(t)\} $ randomly generated from a trajectory pool $\left[\boldsymbol{\tau}^*\right]$, including periodic straight line, squares, circles and figure-eight trajectories. The trajectory pool and trajectory definitions are illustrated in Fig. \ref{fig:trajpool}. This trajectory pool provides domain randomization for different flight missions. The formation controller aims to maintain a desired formation $\mathcal{F}^* = \{\boldsymbol{p}^*_{1i} \mid i \in F\} \in \mathbb{R}^{(N-1)\times 3}$ with control law (\ref{eq:ua}). In the adversarial environment, the attacked agent $a$ aims to learn a dual-mode control policy $\boldsymbol{u}_a \sim \pi(\boldsymbol{u}_a \mid \boldsymbol{s}_a, \{\boldsymbol{s}_j\}_{j \in \mathcal{N}_a})$ to achieve resilient formation no matter if DoS attack occurs. The attacker is fixed at a position, where the periodic DoS attack is launched according to the distance and the attack range in (\ref{eq:dos}).

\subsubsection{State and Observation Space}

The state space for drone \(i\) is defined as \(\boldsymbol{S}_i = [\boldsymbol{s}_i, \{\boldsymbol{s}_j\}_{j \in \mathcal{N}_i}]\), aggregating its own ego-state and the states of its neighbors. The scope of neighbors is limited by the communication range $d_c$. To adopt the dual-mode formation control, the observation space $O$ for the control policy consists of the normal observation $o^1_i \in \mathbb{R}^6$ from the ego-velocity $\dot{\boldsymbol{p}}_i(t)\in \mathbb{R}^3$ and the relative position communicated to the leader ${\boldsymbol{p}}_{1i}(t) \in \mathbb{R}^3$, and ego-centric observations $o^2_i \in \mathbb{R}^{[\mathcal{N}_i]\times 6}$ from neighbors, i.e., relative position and velocity measurements $\{\boldsymbol{p}_{ji}(t), \dot{\boldsymbol{p}}_{j}(t)\}_{j \in \mathcal{N}_i}$. To address the variable number of neighbors, the ego-centric observations are padded into a maximum number $N_{max}$ of observation buffer for the GAT. Therefore, the padded ego-centric observation is $o'_i \in \mathbb{R}^{N_{max}\times 6}$. The normal observation $o^1_i$ is then concatenated with the encoded observation $\boldsymbol{s}'_i$ and fed into the policy NN. When the agent is within the attack range, the communication from the leader is lost, and $\boldsymbol{p}_{1i}(t)$ is set to $[0,0,0]$; otherwise, the ego-centric observations are padded with zeros.

\subsubsection{Action Space}
To align with the Robot Operating System (ROS) command topic, the action for drone \(i\) is defined as the control input \(\boldsymbol{u}_i = [v_{i,x}, v_{i,y}, v_{i,z}]\) to adjust its body velocity, with the allowed range $v_i \in (-v_{min}, v_{max})$. The velocity action can be easily deployed in Sim2Real since the flight controller also uses the velocity commands precisely tracked by corresponding thrust commands in high bandwidth. Since the quadrotor drone is an underactuated system with only the yaw angle individually controlled, we maintain the yaw angle using the default PID control within the flight controller. 

\subsubsection{Reward Function}
In this work, we consider the formation error, power cost and the collision avoidance for the formation control problem simultaneously. The compound reward function \(r_i\) for drone \(i\) at time step \(t\) is designed as:
\begin{equation}
    r_i(t) = -\left(\|\boldsymbol{e}_i(t)\|^2 + \lambda_1 \|(\boldsymbol{u}_i(t)\| + \lambda_2 c(t) \right)/T_{max} \label{eq:reward}
\end{equation}
where \(\boldsymbol{e}_i(t)\) is the relative position error with respect to the desired formation defined in (\ref{eq:formationobj}), which is included to encourage formation convergence even under attack. The item \(\lambda_1 \|(\boldsymbol{u}_i(t)\|\) aims to reduce the power cost brought by the motors. The item $\lambda_2 c(t)$ is to encourage collision avoidance, where $c(t)$ is the collision flag ($c(t)=1$ if there is a collision else 0). \(\lambda_1\) and $\lambda_2$ are the adjustable factors to balance rewards. $T_{max}$ is the allowable maximum number of steps in one episode. 
\begin{algorithm}[!tb]
{\color{black}
\caption{Resilient Formation Control with GAT}
\label{alg::gat}
\textbf{Initialization:}\\ 
{Initialize all agents' states $\boldsymbol{p}_{i}(0)$, $\dot{\boldsymbol{p}}_i(0) = [0,0,0]$, $c(0)=0$, $\boldsymbol{p}_{DoS}$, $episodeCount=0$;} \\
{Initialize training parameters and NNs, including the critic NN $V_{\phi}^0$, the policy NN $\pi_{\theta}^0$ and the GAT encoder NN $GAT^0_{\psi}$, with predefined configurations;}\\
\textbf{Policy Training:}\\
\For{episode ~$= 0$ : maximum episodes}{
{On each episode begins:}\\
Spawn the drones randomly within the 3D space and randomly select one trajectory $\boldsymbol{\tau}^*$ for the leader to track from the trajectory poll.\\
\For{step $t=0:T$}{
\If{The drone $a$ is under attack}{Set $\boldsymbol{p}_{1a}(t) = [0,0,0]$;\\
Collect observations $o^1$ from ego-state and leader state;\\
Collect observations $o^2$ from neighbor state and pad into inputs of the $GAT^0_{\psi}$.}
\Else{Collect observations $o^1$ from ego-state and leader state;\\
Set the inputs of the $GAT^t_{\psi}$ to zeros.}
Execute the formation control policy $\pi_{\theta}^t$ with concatenated observations. Update $c(t)$ and collect mini-batch $\{s(t), a(t), r(t), s(t+1)\}$;
Update the $V_{\phi}^t$;
Conduct policy improvement for $GAT^t_{\psi}$ and $\pi_{\theta}^t$ simultaneously.\\
}}
}
\KwOut{Save the NNs $GAT^t_{\psi}$ and $\pi_{\theta}^t$.}
\end{algorithm}

\subsubsection{Training Schema}
During training, we adopt an actor-critic reinforcement learning algorithm to update the resilient control policy by maximizing the cumulative reward \(R_i = \sum_{t=0}^{T} \gamma^t r_i(t)\), where \(\gamma\) is the discount factor, ensuring that immediate rewards are prioritized over distant rewards, and \(r_i(t)\) is the reward at time step \(t\) in (\ref{eq:reward}). The objective function \(J(\theta)\) for the policy parameters \(\theta\) is defined as the expected value of the cumulative reward:
\begin{equation}
    J(\theta) = \mathbb{E}_{\tau\sim \pi^\theta}\left[R_i\right]
\end{equation}
where the expectation is taken over the distribution of state-action trajectories $\tau =\{s_i(0), a_i(0), s_i(1), a_i(1),\cdots, s_i(T), a_i(T)\}$ induced by the policy $\pi^{\theta}$ parameterized by \(\theta\). The update at iteration \(k\) for a parameter \(\theta\) can be expressed as:
\begin{equation}
    \theta^{(k+1)} = \theta^{(k)} + \alpha \nabla_\theta J(\theta^{(k)})
\end{equation}
where \(\alpha\) is the learning rate and \(J(\theta)\) is the expected return under policy parameterized by \(\theta\). 

In the actor-critic framework, the actor network, parameterized by \(\theta\), defines the policy \(\pi_\theta(a_i|S_i)\), which is the probability of taking action \(a_i\) in state \(S_i\). Here, the actor network is determined by $\pi_\theta(a_i|S_i) = MLP(\boldsymbol{s_i}\| \boldsymbol{s}_i')$, where $MLP(\cdot)$ is a multi-layer perceptron activated by a ReLU and a tanh function. Note that $\boldsymbol{s}_i' = GAT_{\psi}(\boldsymbol{s}_j\mid j \in \mathcal{N}_i)$ is retrieved from the states of neighbours with the aforementioned GAT encoder. The critic network, parameterized by \(\phi\), estimates the value function \(V_\phi(S_i, t) = \mathbb{E}_{\pi^\theta}\left[\sum_{k=0}^{T-t-1} \gamma^k r_i(t+k+1) \mid S_i \right]\), which predicts the expected return from state \(S_i\). The critic helps in evaluating the policy's performance and guiding the actor's updates.

In this work, we adopt PPO algorithm in Unity ML-Agents \cite{juliani2020} to train the actor network formulated with GAT. PPO aims to update the policy by maximizing an objective function while keeping the new policy close to the old policy to ensure training stability. The PPO objective function is:
\begin{equation}
    \mathcal{L}^{PPO}(\theta) = \mathbb{E}\left[\min(r_t(\theta) A_t, \text{clip}(r_t(\theta), 1 - \epsilon, 1 + \epsilon) A_t)\right]
\end{equation}
where \(r_t(\theta) = \frac{\pi_\theta(a_i|S_i)}{\pi_{\theta_{\text{old}}}(a_i|S_i)}\) is the probability ratio of the new policy to the old policy, \(A_t\) is the advantage function at time \(t\), and \(\epsilon\) is a hyperparameter that controls the clipping to avoid overly large policy updates. The advantage function \(A_t\) is computed as the difference between the return and the value estimated by the critic:
\begin{equation}
    A_t = R_i + \gamma V_\phi(S_{i}, t+1) - V_\phi(S_{i}, t)
\end{equation}

Both actor and critic networks are updated iteratively using gradient ascent on \(\mathcal{L}^{PPO}(\theta)\) and gradient descent on the critic's loss, respectively. The overall training algorithm is illustrated in Algorithm \ref{alg::gat}.

\section{Experiment and Results}
In this section, we first present the simulation experiments to demonstrate the superior performance of our approach compared to existing formation approaches. Then the real-world deployment is achieved to demonstrate the Sim2Real capability of trained model without fine-tuning.

\begin{table}[!tb]
\centering
\caption{Hyperparameters Used for Training}
\label{drl-param}
{\color{black}
\resizebox{\columnwidth}{!}{
\begin{tabular}{@{}ll|ll@{}}
\hline
\hline
\textbf{Parameter} & \textbf{Value}  &\textbf{Parameter} & \textbf{Value}\\
\hline

Batch size & 1024 & Buffer size & 10240 \\
Learning rate $\eta$ & 0.0003 & Discount factor $\gamma$ & 0.99\\
Optimizer & Adam & Learning rate schedule & linear \\
Checkpoints & 10 & Time horizon & 128\\
\hline
\textbf{PPO} &  & \textbf{SAC}  & \\
Entropy bonus beta & 0.01 &Interpolation factor $\tau$ & 0.005\\
Clip threshold $\epsilon^{ppo}$ & 0.2 & Update steps & 10.0 \\
Regularization factor $\lambda^{ppo}$ & 0.95 &Initial entropy coefficient & 0.5\\ 
Number of epochs & 3 & Replay size & 1000 \\
\hline
\hline
\end{tabular}
}}
\end{table}

\subsection{Settings}
The training and evaluation in simulation are conducted on a high-performance workstation, with specifications listed as follows: a NVIDIA Geforce RTX 3090Ti GPU, an AMD Ryzen9 5900X CPU, and two Corsair 32G/3600MHZ memory cards. We set the maximum episode length to $T_{max} = 1000$ steps and total number of training steps to 3 million steps. The reward weights are $\lambda_1 = 0.1$ and $\lambda_2 = 0.5$, respectively. 

\begin{figure*}[!tb]
    \centering
    \includegraphics[width=\linewidth]{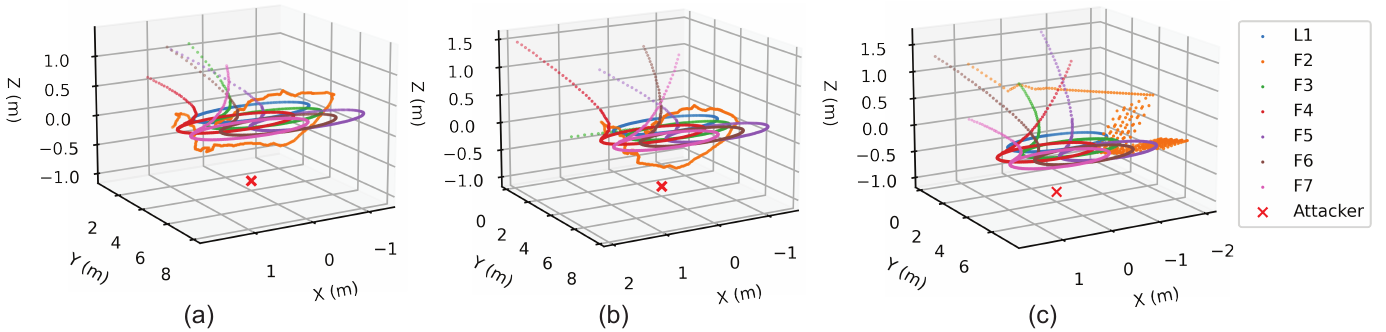}
    \caption{The trajectory of formation in circle under different conditions. (a) GAT-based formation controller under no attack; (b) GAT-based formation controller under DoS attack; (c) displacement-based formation controller under DoS attack. This shows the resilience of GAT-based formation controller.}
    \label{fig:position}
\end{figure*}

\begin{figure}[!tb]
    \centering
    \includegraphics[width=0.8\linewidth]{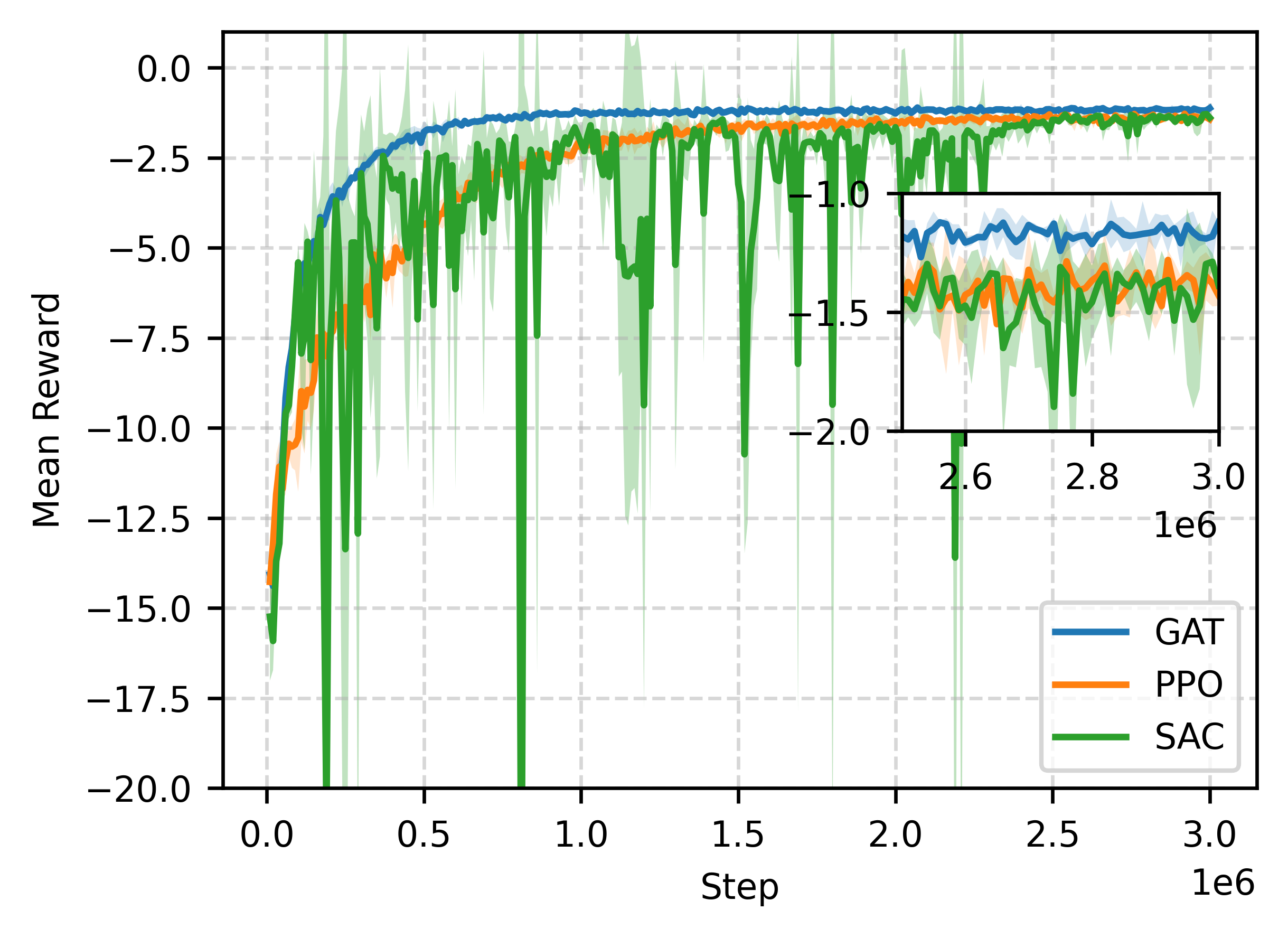}
    \caption{The mean rewards of various DRL approaches. The mean rewards are filled with the standard deviation (Std.) of three training sessions. }
    \label{fig:reward}
\end{figure}

\begin{figure}[!tb]
    \centering
    \includegraphics[width=0.8\linewidth]{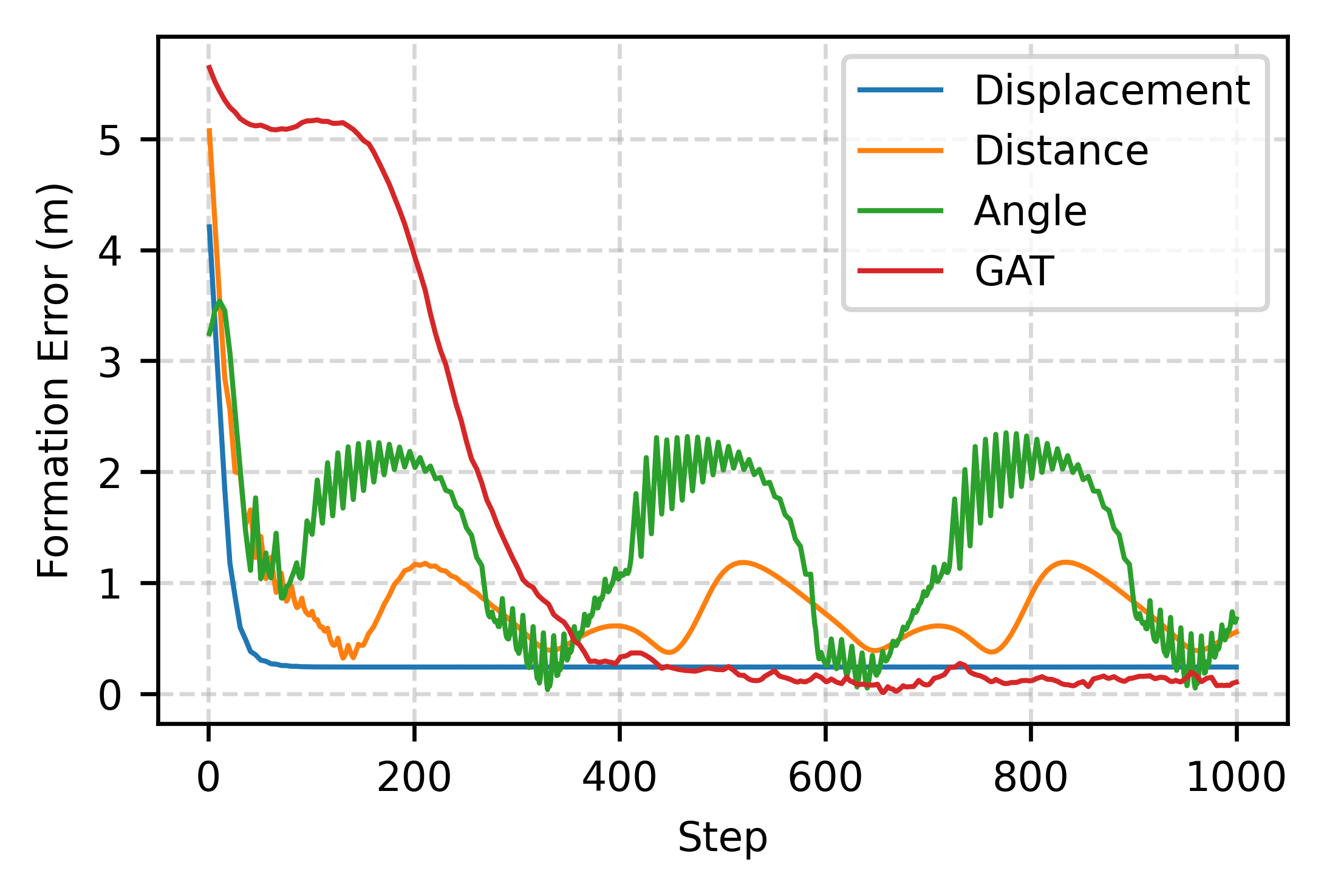}
    \caption{Formation error for various formation controllers under no attack.}
    \label{fig:noattack}
\end{figure}

\begin{figure}[!tb]
    \centering
    \includegraphics[width=0.8\linewidth]{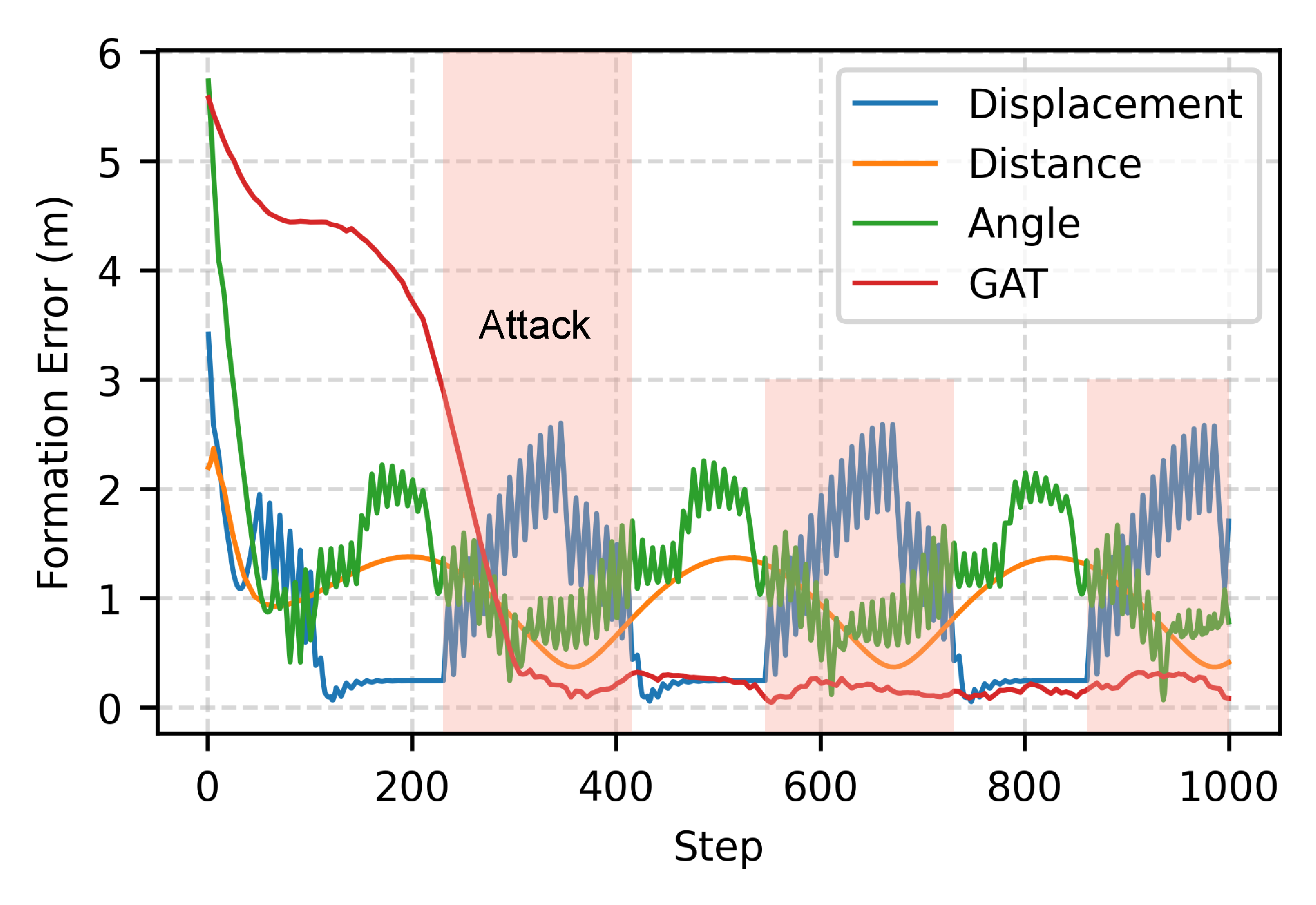}
    \caption{Formation error for various formation controllers under DoS attack.}
    \label{fig:attack}
\end{figure}

\begin{figure}[!tb]
    \centering
    \includegraphics[width=0.8\linewidth]{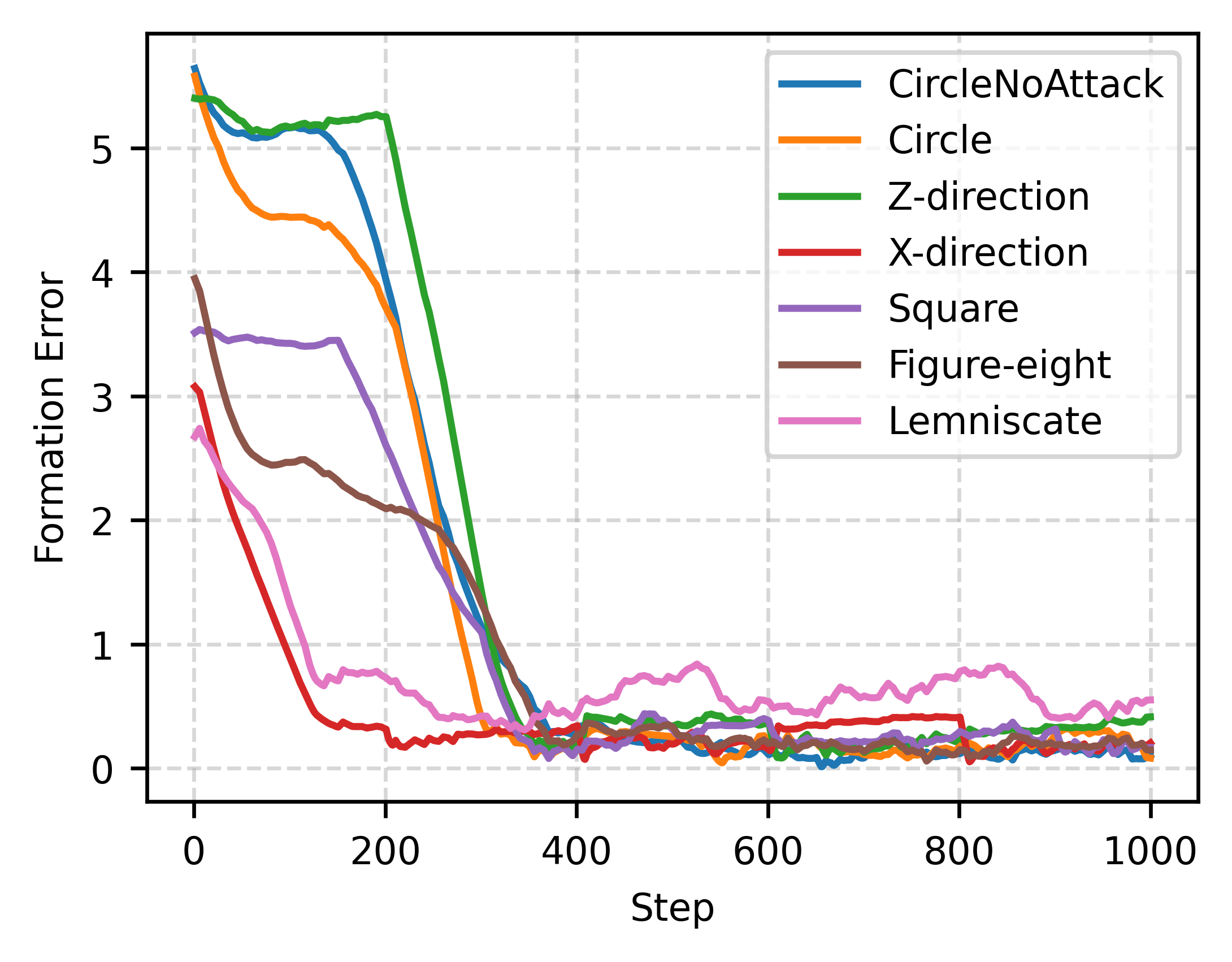}
    \caption{Formation error of GAT-based controller under various missions.}
    \label{fig:formationgat}
\end{figure}

In the training process, we include $7$ agents with 1 leader and 6 followers. The desired formation is defined as $\mathcal{F}^* = \{\boldsymbol{p}^*_{12} = [-0.5, 0, 1], \boldsymbol{p}^*_{13} = [0, 0, 1], \boldsymbol{p}^*_{14} = [0.5, 0, 1], \boldsymbol{p}^*_{15} = [-0.5, 0, 2], \boldsymbol{p}^*_{16} = [0, 0, 2], \boldsymbol{p}^*_{17} = [0.5, 0, 2]\}$ as illustrated in Fig. \ref{fig:sim}. The communication range is $d_c = 0.8m$ (neighbor determination) and the attack range is $\kappa = 3.0m$. The collision threshold is set to $0.15m$. The attacker position is set to $\boldsymbol{p}_{DoS} = [0, -1, 4]$. If the agent is within the attack range, the relative position information to the leader is $\boldsymbol{p}_{1a} = \boldsymbol{0}$. During the training and testing, the leader agent is spawned from a random 3D space: $[Random()*2.0 - 1.2, 3.0 - Random()*0.5, Random()*2.2 - 0.2]$ and the followers are spawned from $[Random()*3.2 - 1.2, Random()*2.2-0.5, Random()*3.2 - 0.2]$, where $Random()$ is a random value uni-distributed over $[0,1]$. The leader tracks the desired trajectories as shown in Fig. \ref{fig:trajpool} while the attacked follower is driven by the policy $\boldsymbol{u}_a$. The actor network and critic network consist of a two-layer fully connected (FC) neural network with 256 nodes in each layer. The output layer of the actor network is 4 node, while that of the critic network is 1 node. For each complete training, we set the number of checkpoints to 10.

\subsection{Metrics and Comparison}
\subsubsection{Metrics} There are two metrics used to evaluate the performance of various methods, namely \textit{formation error} $e_f$ (the main metric) and \textit{collision rate} $cr$, as defined below.
\begin{equation}
        e_f = \frac{\sum_{t=1}^{T_{max}}||\boldsymbol{p}_i(t) - \boldsymbol{p}^*_i(t)||}{T_{max}}
\end{equation}
\begin{equation}
    cr = \frac{number~of~collisions}{T_{max}}
\end{equation}
These two metrics are used to evaluate the formation performance and obstacle avoidance capability, respectively.

\begin{table*}[!tb]
    \centering
    \caption{Formation Performance of Different Formation Controllers under Different Missions ($e_f$($\downarrow$)/$cr$($\downarrow$))}
    \label{tab:collision_error}
    \begin{tabular}{cccccccc}
        \toprule
        \textbf{Mission}& Distance \cite{baillieul2003information} & Displacement\cite{ren2007distributed} & Angle\cite{10040975} & PPO\cite{liu2023distributed} & SAC\cite{li2024distributed} & \textbf{GAT (ours)} \\
        \midrule
        CircleNoAttack                      & 0.8042 / 0.1215  & 0.2934 / 0.0004  & 1.4033 / 0.0559  & 0.4797 / 0.0000 & 1.0180 / 0.0005 & \textbf{0.1529} / 0.0000 \\
        Circle                              & 1.0148 / 0.0608  & 1.7483 / 0.0057  & 1.137 / 0.3870  &  0.2143 / 0.0001 & 2.7360 / 0.0000 & \textbf{0.1977} / 0.0061\\
        Z-direction                         & 1.0046 / 0.0001  & 2.0654 / 0.0016 & 1.3063 / 0.3080 & 0.9405 / 0.0000 & 5.3122 / 0.0000 & \textbf{0.2968} / 0.0010 \\
        X-direction                         & 0.8599 / 0.0890  & 1.2379 / 0.0257  & 0.9931 / 0.3041  & 2.9269 / 0.0000 & 4.4917 / 0.0022 & \textbf{0.2495} / 0.0230 \\
        Square                              & 0.9893 / 0.0395  & 1.5425 / 0.0091  & 1.1410 / 0.0091 & 1.1098 / 0.0000 & 1.4220 / 0.0009 & \textbf{0.2550} / 0.0143 \\
        Figure-eight                        & 0.9069 / 0.0244  & 2.1637 / 0.0026  & 1.1317 / 0.2745  & 1.7055 / 0.0000 & 2.6072 / 0.0001 & \textbf{0.2089} / 0.0003 \\
        Lemniscate & 0.8899 / 0.0101  & 1.6437 / 0.0003  & 1.1497 / 0.4072  & 1.5977 / 0.0005& 3.7323 / 0.0001 & \textbf{0.6018} / 0.0016 \\
        
        \bottomrule
    \end{tabular}
\end{table*}

\subsubsection{Baselines}In this work, we compare our GAT-based formation controller with baseline formation controllers \cite{oh2015survey}, including: 
\begin{itemize}
    \item \textit{Distance-based formation controller} \cite{baillieul2003information}: In distance-based formation control, each agent adjusts its position to maintain a specific distance from the leader. The formation is maintained by distance measurement and the control law is $\boldsymbol{u}_i = k_p\sum_{\mathcal{N}_i}(\|\boldsymbol{p}_j - \boldsymbol{p}_i\| - \|\boldsymbol{p}_{ij}\|^* )\frac{\boldsymbol{p}_j - \boldsymbol{p}_i}{\|\boldsymbol{p}_j - \boldsymbol{p}_i\|}$.
    \item \textit{Displacement-based formation controller} \cite{ren2007distributed}: In displacement-based formation control, each agent adjusts its position to achieve a specific relative position with respect to the leader. The formation is achieved by relative position measurement and the control law is $\boldsymbol{u}_i = k_p(\boldsymbol{p}_1 - \boldsymbol{p}_i - \boldsymbol{p}_{i1}^* )$.
    \item \textit{Angle-based formation controller} \cite{10040975}: In angle-based formation control, each agent adjusts its position to maintain a specific angle with respect to the leader and its neighbors. Note that to use angle-based formation controller, at least two neighbors are considered for 2D and 3 neighbors are considered for 3D space. We assume that agent \( i \) has two neighbors \( j \) and \( k \), and the control law is \[
\boldsymbol{u}_i = k_p \left( \theta_{jik} - \theta_{jik}^{*} \right) \left( \frac{\boldsymbol{p}_j - \boldsymbol{p}_i}{\|\boldsymbol{p}_j - \boldsymbol{p}_i\|} + \frac{\boldsymbol{p}_k - \boldsymbol{p}_i}{\|\boldsymbol{p}_k - \boldsymbol{p}_i\|} \right)
\] where \( \theta_{jik} = \arccos \left( (\boldsymbol{p}_j - \boldsymbol{p}_i) \cdot (\boldsymbol{p}_k - \boldsymbol{p}_i) \right) \) if $(\boldsymbol{p}_j - \boldsymbol{p}_i) \times (\boldsymbol{p}_k - \boldsymbol{p}_i) > 0$ with $\times$ the cross product, and \( \theta_{jik}^{*} \) is the desired angle. Here, we use \( \theta_{123}^{*} = 1.107rad \).
\end{itemize} 
where \( k_p \) is a positive gain, and we set $k_p=6.0$ for all controllers. To ease the oscillation, a damping term $-k_v\dot{\boldsymbol{p}}_1$ with $k_v = 0.5$ is added to the control law as expressed in (\ref{eq:controller3}). We also compare our method with two other DRL-based approaches, namely PPO \cite{liu2023distributed} and SAC \cite{li2024distributed}. The hyperparameters of the training are listed in Table \ref{drl-param}.

\subsection{Results and Discussion}

The training curves of different DRL approaches are illustrated in Fig. \ref{fig:reward}, demonstrating the effectiveness of GAT over other RL approaches. GAT outperforms PPO and SAC in terms of sample efficiency and final rewards, while SAC shows an unstable training process in dynamic environments. We first test our trained control policy in a simulation of an open field with different leader trajectories, including all the aforementioned trajectories in Fig. \ref{fig:trajpool}. An unseen complex 3D Lemniscate trajectory \cite{lawrence2013catalog} with $[x,y,z] = [\frac{1.2\cos(t)}{1 + \sin^2(t)}, 1.2\cos(t), \frac{1.2 \cos(t) \cdot \sin(t)}{1 + \sin^2(t)}]$ is added to test the generalization capability of the trained control policy. The leader's initial position is from $[0,0,4] + RandomSphere(0.5)$, where $RandomSphere(0.5)$ represents a uniform distribution within a sphere of radius 0.5. Over 10 episodes, we calculate the average collision rate and formation error. The performance results are listed in Table \ref{tab:collision_error}. The formation errors of various formation controllers under no attack and under DoS attack in the Circle mission are illustrated in Fig. \ref{fig:noattack} and Fig. \ref{fig:attack}, respectively.

From Table \ref{tab:collision_error}, it is evident that the GAT-based formation controller achieves the lowest formation error across all missions, regardless of DoS attacks. As shown in Fig. \ref{fig:noattack}, the formation converges with the lowest formation error under no attack. The periodic DoS attack in the Circle mission affects all formation controllers, but the GAT-based formation controller demonstrates the capability to mitigate the oscillations, as illustrated in Fig. \ref{fig:attack}. It is important to note that the control power cost is considered in the reward function (\ref{eq:reward}), which leads to a longer response time for the GAT-based controller compared to the displacement-based method, resulting in a slightly higher formation error at the beginning. The formation errors of GAT-based controller for all missions are illustrated in Fig. \ref{fig:formationgat} and the trajectories of formation in Circle mission under several conditions are illustrated in Fig. \ref{fig:position}, which show the resilience and robustness of proposed GAT-based formation control.

\begin{figure}[!tb]
    \centering
    \includegraphics[width=\linewidth]{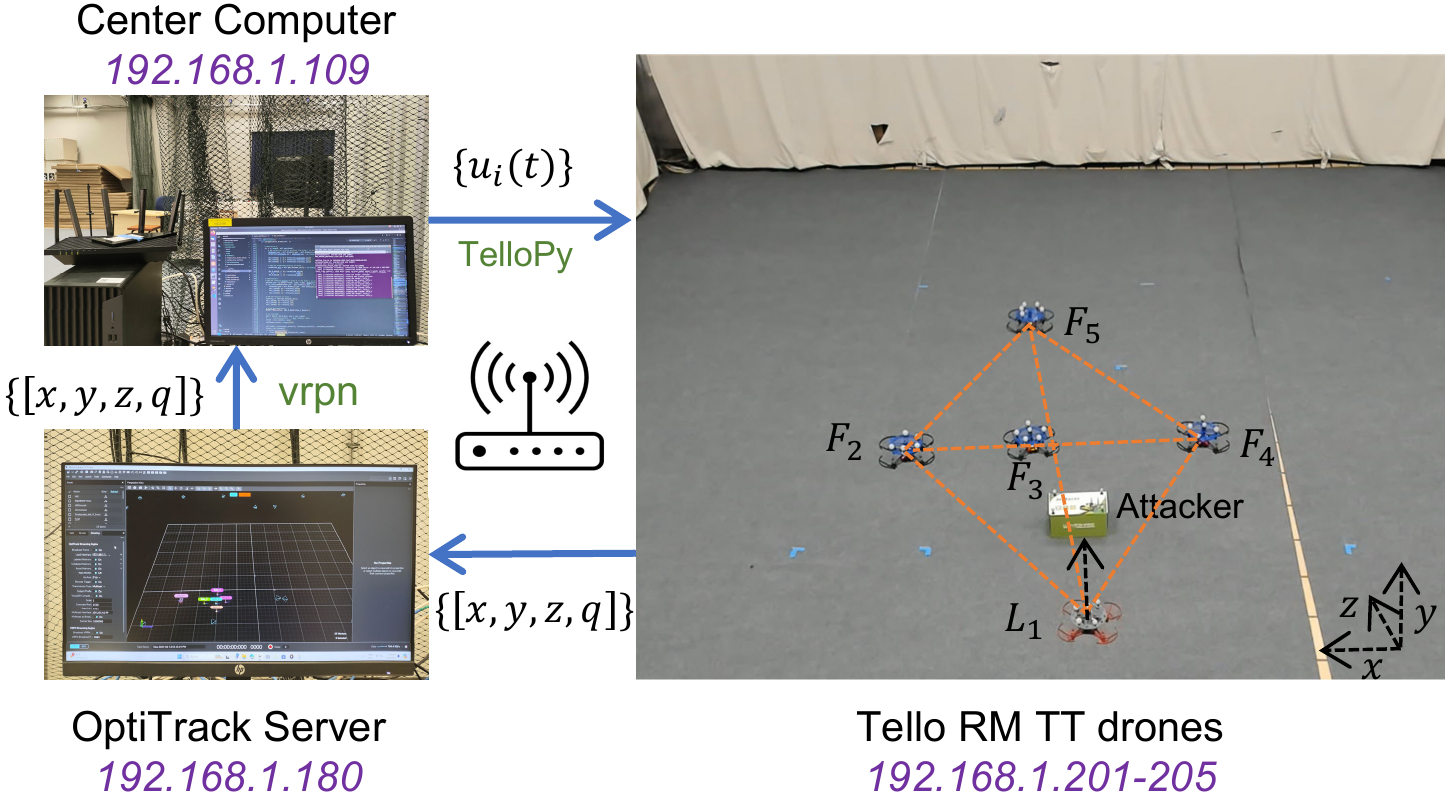}
    \caption{Physical experiment configurations where all drones, OptiTracker motion capture system and center computer are connected to the same local network with different IP addresses.}
    \label{fig:phyexp}
\end{figure}

\begin{figure*}[tb]
    \centering
    \includegraphics[width=0.95\linewidth]{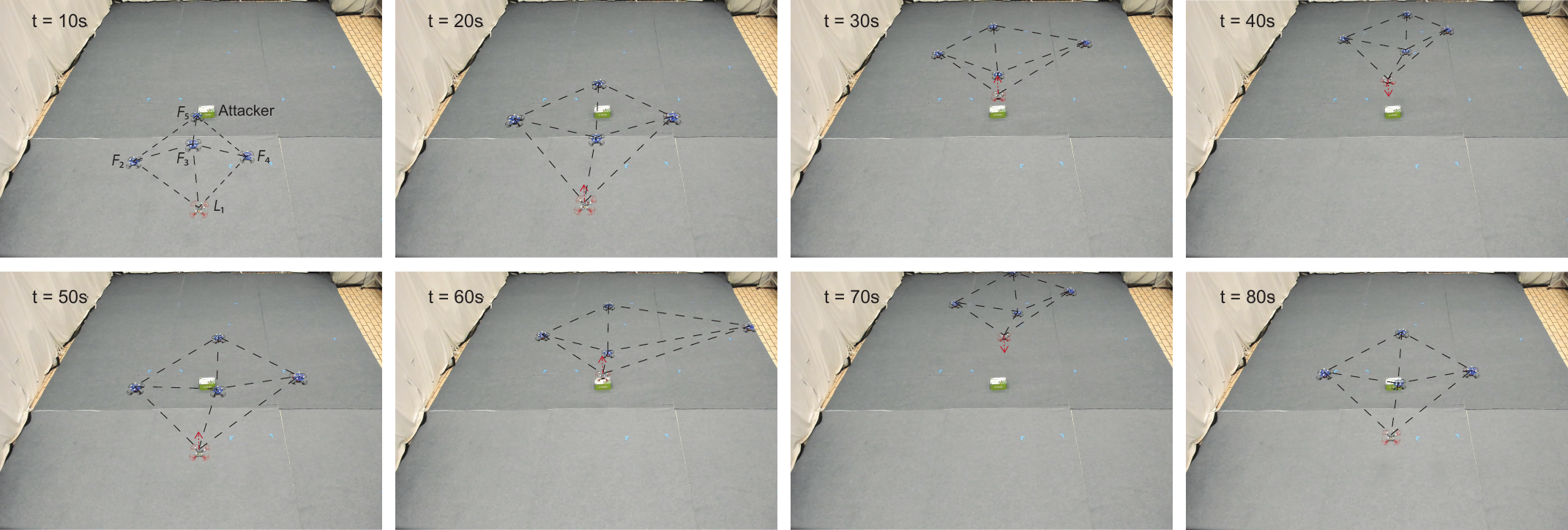}
    \caption{Snapshots of the GAT-based formation maneuvering in the real-world flight with zero-shot Sim2Real (red arrow is the moving direction).}
    \label{fig:snapshot}
\end{figure*}

\begin{figure}[tb]
    \centering
    \includegraphics[width=\linewidth]{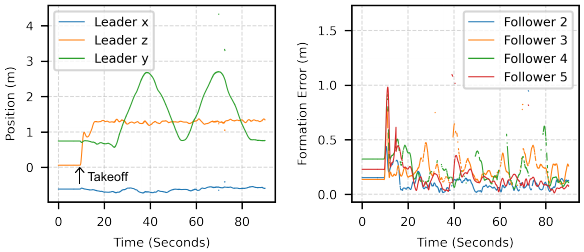}
    \caption{The position of the leader (left) and the formation error of each follower (right) in the Sim2Real physical experiment. Note that the data outliers are caused by the unstable positioning of MCS.}
    \label{fig:simrealerror}
\end{figure}

\subsection{Real-World Deployment}
We further conduct real-world deployment to validate the effectiveness of trained policies with zero-shot Sim2Real. The desired formation is not trained in the simulation with relative positions as $\{ \boldsymbol{p}^*_{12} = [-0.8, 0.8, 0], \boldsymbol{p}^*_{13} = [0, 0.8, 0], \boldsymbol{p}^*_{14} = [0.8, 0.8, 0], \boldsymbol{p}^*_{15} = [0, 1.6, 0] \}$. We use five Tello RM TT drones (the detailed specifications can refer to \url{https://www.dji.com/sg/robomaster-tt/specs}) for physical experiments to test the generalization capability of the trained model. The attacker is simulated with a green box with the position $\boldsymbol{p}_{DoS}=[-0.738, 2.915, 0.102]m$. Each Tello drone is set to AP mode with a distinct fixed IP address from 192.168.1.201 to 192.168.1.205. The center computer is running with Ubuntu 18.04 and the CPU Intel Core i7. Within the same local network, the physical experiment connections are illustrated in Fig. \ref{fig:phyexp}. All drones are attached with reflective ball markers for positioning with the connected OptiTrack motion capture system (MCS), which broadcasts the position information at a frequency of 120 Hz. Note that the velocity observation are set as the velocity commands from last time step. All drones are controlled from the center computer side via the TelloPy ROS package with distributed ROS nodes. The trained model is loaded in ONNX format and executed on the center computer side. All measures are processed to align with the inputs of the trained model. As in the simulation setting, the position values are set to zeros if the agents are within the attack range of the attacker (the green box), i.e., $3.0m$. The leader follows a forward-backward motion along the y-direction with distance of $2.0m$. The snapshots of the formation maneuvering of five drones are illustrated in Fig. \ref{fig:snapshot} and the relative position errors for each drone under attack are plotted in Fig. \ref{fig:simrealerror}, which show the resilience of the GAT-based formation controller in the Sim2Real deployment.

\section{Conclusion}
Resilient formation control is imperative for multidrone systems operating in unstable communication and adversarial environments with potential cyberattacks. Compared to traditional formation controllers designed for specific conditions, this paper proposes a GAT-based formation controller that can handle communication loss by extracting state features from available neighbors using attention scores. Extensive simulations demonstrate that our controller outperforms traditional formation controllers under both no-attack and DoS attack scenarios. Furthermore, real-world flight tests with zero-shot Sim2Real validate the effectiveness of the trained model.

\bibliographystyle{IEEEtran}
\bibliography{IEEEabrv,ref}

\vskip -1.5\baselineskip plus -2fil

\end{document}